\documentclass[ChapterTOCs,nemilov1,sectionauthor]{Nemilov}

\usepackage[english]{babel}
\usepackage{amssymb}
\usepackage{amsmath}
\usepackage{graphicx}
\usepackage{subfigure}
\usepackage{makeidx}
\usepackage{multicol}
\usepackage[utf8]{inputenc} 
\usepackage{epigrafica}
\usepackage[LGR,OT1]{fontenc}
\usepackage{mathptmx}
\usepackage{booktabs}
\usepackage{url}
{\par\addvspace{6pt}\normalfont {\bfseries #1}\hskip\labelsep\ignorespaces\itshape}
{\par\addvspace{6pt}}

\AtBeginDocument{\renewcommand\partname{Section}}  

\makeindex

\begin{document}

\title{ Decisions and Deployment: The Five-Year SAHELI Project (2020–2025) on Restless Multi-Armed Bandits for Improving Maternal and Child Health Behaviors}

\author{Shresth Verma $^1$, Arpan Dasgupta $^2$, Neha Madhiwalla $^3$, Aparna Taneja $^2$, Milind Tambe $^1$\\[15pt] \AffiliationFont $^1$ Harvard University, USA \\[2pt]
\AffiliationFont $^2$ Google Deepmind, India \\
\AffiliationFont $^3$ ARMMAN, India}


\maketitle


\setcounter{chapter}{4} 
\chapter[ Restless Multi-Armed Bandits for Maternal and Child Health]{Restless Multi-Armed Bandits for Maternal and Child Health}
\markboth{RMAB for Maternal Health}{Verma et al.: SAHELI Project}
Global health initiatives often face a fundamental challenge: how to allocate scarce financial and human resources to effectively reach vulnerable populations which are most in need \cite{garrett2017challenge}. This challenge is particularly important in maternal and child health, where timely information and care can significantly influence health outcomes. Countries in the Global South, including India, continue to experience high maternal and neonatal mortality rates, which are partly attributed to a lack of access to reliable health information and services \cite{khalil2023call}. Reducing the information gap in such scenarios can therefore have a huge impact.

The widespread penetration of mobile phones, even in low-resource settings, has unlocked the potential for Mobile Health (mHealth) programs to deliver preventive health information directly to underserved communities \cite{stanton2016mobile,early2019use,armman-mhealth, madanian2019mhealth}. NGOs and public health organisations frequently use automated voice message services to support pregnant women and new mothers with guidance. However, despite high adoption, maintaining beneficiaries' engagement remains a critical challenge in such public health information programs \cite{armman-mhealth, info:doi/10.2196/35371, info:doi/10.2196/jmir.7.1.e11, info:doi/10.2196/20283}. 
Beneficiaries often have a limited understanding of the importance of continued listenership or may have logistical issues such as inconsistent access to mobile phones and networks, leading to declining engagement and high drop-off rates.

To counter this, several non-profits take help from a support centre from which trained healthcare workers can deliver live service calls to beneficiaries. These calls represent a human touch in digital health which can encourage participation, troubleshoot logistical issues, and offer personalised reassurance. However, because the number of beneficiaries vastly exceeds the available health workforce, identifying which beneficiaries should receive these scarce live calls is a critical optimization problem. 

Adherence to healthcare programs is also a well-studied problem in other domains such as tuberculosis~ \cite{Killian_2019,  10.1001/archinte.1996.00440020063008}, cardiac problems~ \cite{son2010application, corotto2013heart}, elderly care  \cite{pollack2002pearl}, HIV  \cite{HIV},  lifestyle changes  \cite{liao2020personalized}, and more. These works either use single-shot prediction of beneficiaires with high risk of drop-out or perform decision making sequentially to choose a small set of beneficiaires for intervention so as to maximize adherence. 
Restless Multi-Armed Bandit (RMAB) is a popular framework in multiagent systems to model the latter problem of sequential allocation of limited resources  \cite{whittle1988restless,nino2023markovian}.
RMABs have been used in various domains of multiagent problems such as machine maintenance  \cite{DBLP:journals/informs/AbbouM19}, anti-poaching  \cite{DBLP:conf/atal/QianZKT16}, and healthcare  \cite{ayer2019prioritizing,verma2022saheli}. 

This chapter presents SAHELI, an AI-driven system developed to address these challenges in partnership with ARMMAN \cite{armman}, one of the world’s largest maternal and child health mHealth NGOs. SAHELI (System for Allocating Healthcare-resources Efficiently under Limited Interventions), meaning “female friend” in Hindi, is designed to assist health workers by smartly scheduling limited service calls. Using a Restless Multi-Armed Bandit (RMAB) framework, SAHELI predicts the beneficiaries' listenership patterns and outputs a live-service call scheduling policy. 

\textit{SAHELI operated in actual deployment from April of 2022 and benefited over 350,000+ mothers.} This book chapter summarizes findings from this five year SAHELI project that began in Spring of 2020, and lasted till summer of 2025, in partnernship between AI researchers and ARMMAN, reported in a number of publications \cite{verma2022saheli, verma2023decision, verma2023increasing, mate2022field, verma2023restless, wang2023scalable, wang2021learning, verma2024leveraging}.
We move from theoretical formulation to practical deployment, detailing innovations such as predicting beneficiary behaviour and shifting from a predict-then-optimize paradigm to decision-focused learning. Finally, we present evidence from large-scale field studies demonstrating not only sustained improvements in beneficiary engagement but also statistically significant positive shifts in key maternal health behaviours. These results show that AI-augmented outreach can meaningfully strengthen maternal and child health programs, offering a blueprint for responsible and effective resource allocation in constrained settings worldwide. 

Key takeaways from this research include: (i) Restless bandits are a very promising approach that provided real world benefits by optimizing limited public health interventions; (ii) Decision-focused learning is an important technique in predict-then-optimize settings; (iii) Results of long term intervention studies showed that restless bandits not only improved adherence to mHealth programs, but ultimately led to improved health behaviors.

\section{Mobile Health Context}[subsection]

To understand the application of AI in mobile health interventions, we focus on a specific case study in collaboration with ARMMAN. ARMMAN is a non-governmental nonprofit organisationn in India dedicated to improving maternal and child health outcomes among underserved and disadvantaged communities. Their flagship initiative, mMitra, is an mHealth service that leverages the extensive penetration of mobile phones in India to deliver critical preventive health information to expectant and new mothers via automated voice messages. As of 2025, 82.74\% of India’s population had access to wireless telecom services, reflecting the broad cellular coverage that makes such mobile outreach feasible \cite{TRAI2025TelecomIndicators}. A significant proportion of beneficiaries, approximately 90\%, live below the World Bank international poverty line. Despite the economic challenges these mothers face, automated voice messaging has proven to be a feasible and scalable mechanism for disseminating vital health information, thanks to the widespread availability of low-cost phones \cite{pfammatter2016mhealth, Kaur2020smart}.

Upon enrolment in the mMitra program, beneficiaries receive brief 1–2 minute messages tailored to their gestational age or the age of their infant. While this approach has been effective in improving maternal health knowledge and behaviors, engagement often declines over time, with roughly 22\% of beneficiaries dropping out within the first three months.

To mitigate disengagement, ARMMAN supplements automated messages with live service calls from trained health workers. These calls, made to a limited number of beneficiaries weekly, help encourage participation, address questions or requests, and prevent further dropouts. This raises a key operational question: which beneficiaries should be prioritized for live service calls to maximize engagement and improve maternal health outcomes?

\subsection{Challenges in service call planning}[subsection]
The challenge of optimising who to reach is a sequential resource allocation problem. The NGO must continuously make the decision about where to invest health workers' time and effort today to achieve the best long-term outcomes. The key challenges in this scenario are that

\begin{itemize}
    \item \textbf{Stochastic and Dynamic State:} A beneficiary's engagement is not static but a stochastic process. Furthermore, a beneficiary's state evolves over time, even if no intervention is made.
    \item \textbf{Resource Constraint:} The weekly budget for service calls is constrained to a fixed number ($K$) across a population of thousands ($N$).
    \item \textbf{Long-Term Impact:} The optimization goal is not just a temporary increase in listenership, but a sustained, long-term improvement in engagement which should lead to positive health behaviors.
\end{itemize}

\section{Methodological Framework}[method]

\subsection{The Restless Multi-Armed Bandit (RMAB) Model}[model]

The \text{Restless Multi-Armed Bandit (RMAB)} framework is a popular framework used to model sequential resource allocation problems under a limited budget. In an RMAB, we have a set of $N$ arms, each modelled as an independent \text{Markov Decision Process (MDP)} defined by the tuple \(\{S, A, R, P\}\). Each of the N arms is in a current state $S_i \in S$ representing its current status. At each decision step, an action \(a_i \in A =\{0,1\}\) is selected for every arm, where \(a_i = 1\) indicates that the arm is actively chosen for intervention and \(a_i = 0\) indicates that it remains passive. The evolution of each arm's state is governed by the transition probabilities \(P(S_{i,t+1}, a_{i,t}, S_{i,t})\), which define the distribution over next states conditioned on the current state and action. The immediate reward obtained from arm \(i\) at time \(t\) is given by \(R(S_{i,t})\). The decision-maker must also satisfy a resource constraint that limits the number of arms that can be activated simultaneously, namely
\[
\sum_{i=1}^N a_{i,t} \le K.
\]

The objective is to find a policy $\pi$ that maximises the expected total discounted reward over a long time horizon. RMAB problems are generally PSPACE-hard to solve optimally. However, Whittle  \cite{whittle1988restless}, proposed an
index-based heuristic that can be solved in
polynomial time and is now the dominant technique used
for solving RMABs.
This method introduces a passive subsidy $\lambda$, which is a virtual reward added to the standard reward when taking the passive action. For each state $s$, the Whittle index is defined as the smallest subsidy that renders the planner indifferent between taking the active and passive actions:
\[
W(s) = \inf_{\lambda} \{ \lambda \,:\, Q_{\lambda}(s,p) = Q_{\lambda}(s,a) \},
\]
where $Q_{\lambda}$ denotes the action-value function under subsidy $\lambda$. This subsidy can be efficiently computed using a binary-search based algorithm. Finally, the index policy chooses K arms with the highest Whittle index for selection at every timestep.


\subsection{Problem Formulation: mHealth Interventions as an RMAB}[problem]

The mHealth service–call allocation problem can be naturally modelled within the RMAB framework. Each beneficiary corresponds to an independent arm whose engagement behaviour is represented by a two–action Markov Decision Process (MDP). Pulling an arm corresponds to taking the active action, i.e., placing a live service call to the beneficiary (denoted by $\alpha = a$), whereas the passive action $\alpha = p$ represents abstaining from a call for that decision epoch. The health–worker team's weekly calling capacity imposes the standard RMAB resource constraint that at most $K$ arms may be activated in any week.

\begin{figure}
    \centering
    \includegraphics[width=0.5\linewidth]{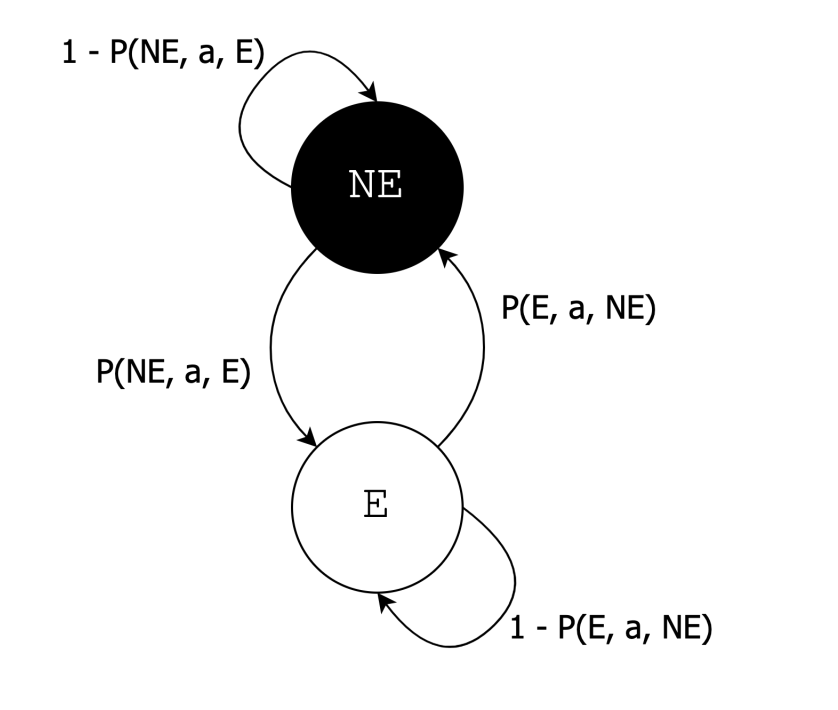}
    \caption{The beneficiary transitions from a current state s
to a next state s'
under action a, with probability P(s, a, s'). Here the two states are \{NE, E\}.}
    \label{fig:MDP}
\end{figure}

The state space associated with each beneficiary is binary-valued and captures their recent engagement behaviour with the automated voice calls. Formally, the state $s \in S = \{0,1\}$ takes the value $1$ if the beneficiary is in the Engaging (E) state and $0$ if they are in the Not Engaging (NE) state. In the operational definition used by ARMMAN, a beneficiary is considered engaging during a week if they listen to at least one automated voice message for more than $30$ seconds (the average message length is one minute). Otherwise, the beneficiary is classified as Not Engaging for that time step. We consider two actions, $\alpha \in A = \{p, a\}$, where the passive action (p) refers to not making a service call, while the active action (a) refers to making a live service call by the health worker. Finally, the state of the beneficiary evolves according to a two–state Markov model with transition probabilities $P(s', \alpha, s)$, where $P(s', \alpha, s)$ denotes the probability of transitioning from state $s$ to state $s'$ when action $\alpha$ is applied.
This is shown in Figure \ref{fig:MDP}.

The reward structure reflects the goal of maximising long-term engagement with the automated voice messages program. We define the reward associated with a beneficiary in state $s \in \{0,1\}$ is given by $R(s) = s$, meaning that an Engaging state yields a reward of $1$ while a Not Engaging state yields $0$. 

To compute a tractable policy for large populations, we use the Whittle index approach for RMABs. 
In this problem, the timestep is defined to be a week, taking into account the NGO's planning pipeline. By modelling the system this way, the Whittle Index $\mathcal{W}(S_i)$ serves as a priority score for each beneficiary. The AI system thus prioritises the $K$ most impactful interventions.


\subsection{Practical Implementation: Learning RMAB Parameters from Data}[lessons]

While the RMAB framework provides the theoretical structure, applying it in practice requires transition probability parameters for every beneficiary (arm) to be known beforehand. Here, we show two approaches for inferring them from historical data and subsequently making the decisions through Whittle-Index policies.

First, we describe the data that is collected as part of the program. When beneficiaries enroll into ARMMAN’s mmobile health program, their registration information is collected which includes age, education level, income bracket, phone ownership, gestational age, number of children, preferred language, and preferred call time slots. These features are referred to as Beneficiary Registration Features and are stored alongside anonymised listenership logs of automated voice messages. This dataset supports the estimation of transition probabilities and the evaluation of engagement dynamics necessary for training and testing RMAB-based policies.

\subsection{Two-Stage Learning}[ts]
In a Two-Stage learning setup, a predictive model is first learnt to infer the transition probability of a beneficiary. Subsequently, the estimated transition probabilities are used to calculate the Whittle Index and then actions are taken based on the Whittle Index policy.
Formally,  the mapping function \( f \) is learned to predict the transition probabilities of each arm i using its socio-demographic features \( x_i \). The predicted transition probabilities for arm \( i \) are then given by
\[
P_i = f(x_i), \qquad i \in [N].
\]
Since our problem domain consists of two states and two actions, the model must predict four transition probabilities. The mapping function can be parameterised as a neural network \( f_w \) with weights \( w \), and a final logistic activation function ensures valid probability values.

The model \( f_w \) is trained by minimising the negative log-likelihood of the observed trajectories \( T \) under the predicted transition probabilities \( f_w(x) \). The loss function \( L \) is defined as
\[
L(f_w(x), T) = \mathbb{E}_{i \in [N]} \big[-\log\big(T_i \mid f_w(x_i)\big)\big],
\tag{3}
\]
and the weights \( w \) of the neural network are optimised via gradient descent by backpropagating
\[
\frac{d\,L(f_w(x), T)}{d w}.
\]

\subsection{Decision-Focused Learning}[dfl]

One downside of Two-stage learning is that there is a mismatch between the objective that is maximised (predictive accuracy in transition probabilities) and the objective that is desired (higher listenership of beneficiaries).
\textit{Decision Focused Learning }fixes this problem by directly optimising the final decision outcome \cite{wilder2019melding, wang2020scalable, wang2021learning, shah2022decision}. 

Off-Policy Policy Evaluation (OPE) is used to quantify this outcome, measuring the reward achieved by a learned policy when applied to past trajectories that were generated by a different policy. 
The DFL architecture uses the same predictive model $f_w$ as TS, described previously. 
After the transition probabilities are predicted as $P = f_w(x)$, we compute Whittle indices through a differentiable function $W$. 
These Whittle indices, $WI = W(P)$, define a differentiable policy denoted by $\pi_{WI}$. 
The final objective is obtained by evaluating this learned policy using OPE on the observed trajectories $T$, expressed as $\text{OPE}(\pi_{WI}, T)$. 
The predictive model parameters are optimised by maximising this objective and backpropagating through the entire pipeline. 
The resulting gradient is
\[
\frac{d\, \text{OPE}(\pi_{WI}, T)}{dw}.
\]

In Decision-Focused Learning, this gradient is computed using the chain rule:
\[
\frac{d\, \text{OPE}(\pi_{WI}, T)}{dw}
=
\frac{d\, \text{OPE}(\pi_{WI}, T)}{d\pi_{WI}}
\cdot
\frac{d\pi_{WI}}{d WI}
\cdot
\frac{d WI}{dP}
\cdot
\frac{dP}{dw}.
\]

\section{Empirical Evaluation: Field Studies with ARMMAN}[ee]




To evaluate the real-world effectiveness of AI AI-based intervention strategy, we conducted a Randomised Controlled Trial (RCTs) in collaboration with ARMMAN. 

\subsection{Study 1: Impact on Program Engagement and Listenership}[s1]

In the pilot study, we evaluate the performance of Two-Stage against Round Robin, a heuristic-based method and Current Standard of Care, which is a baseline where no live service calls are sent out. This study was conducted by 
Mate et al. (2022)  \cite{mate2022field} in April. It spanned 7 weeks and constituted 23,003 beneficiaries who were distributed
in three groups corresponding to the RMAB policy, Round
Robin (RR) and Current Standard of Care (CSOC). The results from this pilot study are shown in Table \ref{tab:2022}.

\begin{table}[h!]
\centering
\begin{tabular}{lcccc}
\toprule
\textbf{} & \textbf{RMAB vs CSOC} & \textbf{RR vs CSOC} & \textbf{RMAB vs RR} \\
\midrule
\% reduction in cumulative engagement drops 
    & 32.0\% & 5.2\% & 28.3\% \\
p-value 
    & 0.044$^{*}$ & 0.740 & 0.098$^{\dagger}$ \\
Coefficient $\beta$ 
    & -0.0819 & -0.0137 & -0.0068 \\
\bottomrule
\end{tabular}
\label{tab:2022}
\caption{Statistical significance for service call policy impact at week 7 is tested using a linear regression model. 
We use: $^{*}p < 0.05$; $^{\dagger}p < 0.1$.}
\end{table}

The pilot results demonstrated that the RMAB method cuts ~ 30\% of the beneficiary engagement drops experienced by the other groups. Furthermore, whereas RMAB
achieves statistically significant improvement against CSOC
(p $<$ 0.05) and RR (p $<$ 0.1), RR fails to achieve any statistical signifacnt advantage over CSOC.

Next, we evaluate the performance of Decision Focused Learning.
While DFL has demonstrated strong performance in simulated domains, evidence of its practical utility in deployed settings remains limited. These trials thus establish whether Decision Focused Learning has statistically significant benefit in listenership over Two-Stage and Control baselines.


We first identified a pool of eligible beneficiaries and randomly sampled 9000 participants. These were then assigned to three equally sized groups of 3000 each: 
\begin{enumerate}
    \item \emph{Current Standard of Care} (CSOC) or control group that did not receive live service calls,
    \item \emph{Two-Stage} (TS) group in which live calls were scheduled according to a Whittle Index policy trained using a two-stage learning pipeline, and
    \item \emph{Decision-Focused Learning} (DFL) group in which the Whittle Index policy was learned end-to-end with the final decision objective.
\end{enumerate}
Additional checks were added to make sure that the socio-demographic features and initial engagement states were balanced across groups.

Participants in the TS and DFL groups were eligible to receive live service calls, whereas the CSOC group served as a baseline for natural engagement levels. Each week of the intervention, we selected $K = 300$ beneficiaries per treatment group for live service calls in accordance with the respective learned scheduling policies. This intervention was carried out over 6 consecutive weeks, during which all groups continued to receive standard automated voice messages. 
This Randomized Control Trial (RCT) design thus tested the relative performance of AI-driven scheduling strategies over the Current Standard of Care and specifically whether DFL provides measurable improvements over the two-stage approach.

The primary metric of success is defined using Engagement metric. The engagement at time \( t \) for the \( i \)-th beneficiary is denoted as \( E_i(t) \). $E_i(t)$ is defined as 1 if the beneficiary listens to at least one automated call in a given week for more than 30 seconds, and 0 otherwise. Since engagement typically declines over time, we measure the reduction in engagement relative to the initial engagement. The engagement drop and the cumulative engagement drop are defined as follows:

\[
E_i^{\text{drop}}(t) = E_i(0) - E_i(t)
\]
\[
E_i^{\text{cumulative\_drop}}(t) = \sum_{\zeta=0}^{t} E_i^{\text{drop}}(\zeta)
\]

\begin{figure}
    \centering
    \includegraphics[width=0.8\linewidth]{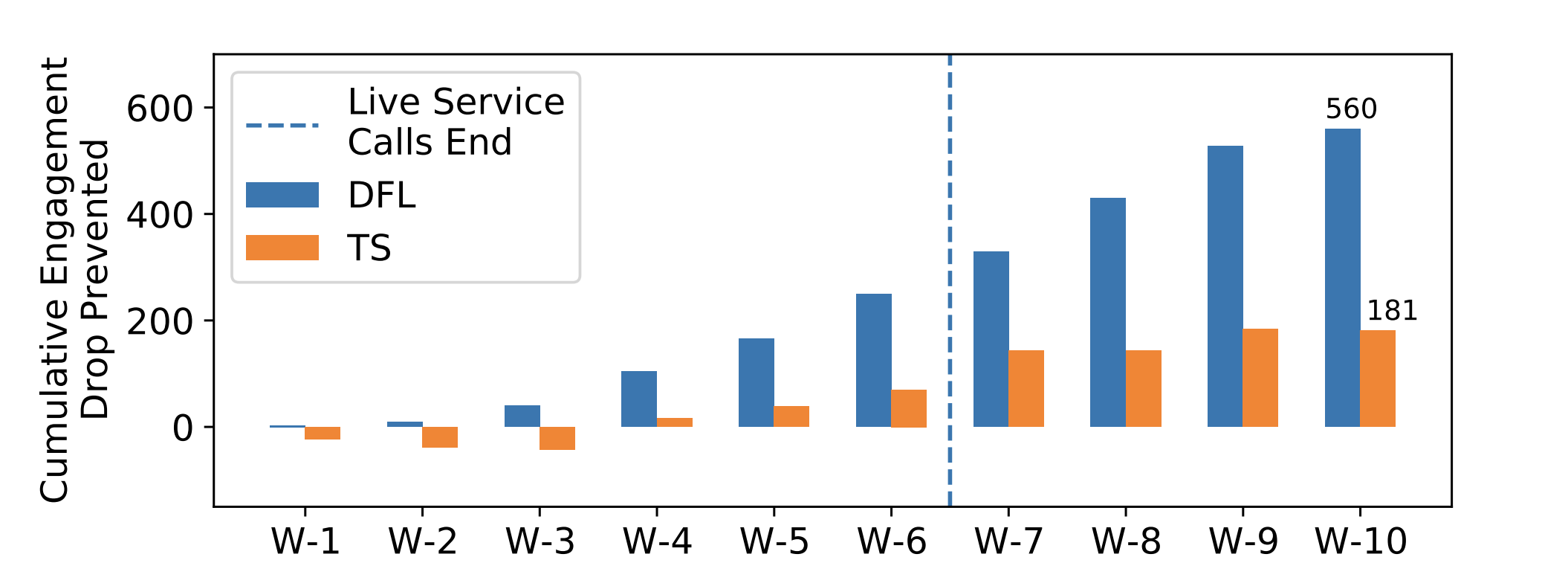}
    \caption{Weekly Cumulative Engagement Drop Prevented
for the DFL and TS groups. Live service calls are only delivered for the first 6 weeks, after which, all three groups
are only passively observed. The DFL group prevents more
Cumulative Engagement Drops as compared to the TS group}
    \label{fig:dflstudy}
\end{figure}
The cumulative engagement drop prevented by the CSOC group is the difference in the cumulative engagement drop between the policy and the CSOC group up to time $t$. Figure \ref{fig:dflstudy} illustrates the cumulative engagement drops prevented over the CSOC group for both DFL and TS policies. We observe that the DFL policy prevents more engagement drops across all weeks. By the end of the study, the DFL group prevented 560 more engagement drops compared to the TS group, which only prevented 181 drops. Given that the CSOC group experiences a total of 1,765 cumulative engagement drops, the DFL group shows a 31\% reduction in cumulative engagement drops relative to the CSOC group, while the TS group results in a 10\% reduction.

\subsection{Study 2: Translating Engagement to Health Behaviour Change}[s2]

While improved listenership is a strong proxy, the ultimate metric is whether the increased exposure to information translates into positive health behaviour change. A subsequent large-scale study was designed to explicitly address this open question.

The study focused on measuring key Postnatal Health behaviours that are strongly correlated with maternal and child well-being, specifically the consumption of Iron and Folic Acid (IFA) supplements and Calcium supplements in the postnatal period.

The study was conducted with multiple cohorts totalling 34453 beneficiaries. Within each cohort, beneficiaries were randomly assigned to either the intervention arm (based on DFL algorithm) or the control arm, while ensuring a similar distribution of key socio-demographic and listenership behaviours. Interventions were given to roughly 35\% of the beneficiaries in the treatment cohort. Finally, while conducting intervention based on DFL in the intervention arm, a matching list of beneficiaries in the control arm selected for intervention by DFL, but not delivered a service call was also maintained. These counterfactual counterparts of the DFL group in the control group are then used to make comparisons in behaviours. 

To measure the impact of AI-based interventions on health knowledge and behaviour of beneficiaries, a computer-assisted telephonic interview (CATI) survey was conducted to interview beneficiaries at the end of three months from their estimated date of delivery. Beneficiaries who were intervened in the DFL group and the similar counterfactual beneficiaries in the control group were called by health workers and asked survey questions across various health topics to gauge the beneficiaries’ comprehension of the delivered
information in automated voice messages. For each question, beneficiaries received a score based on their responses. Further,  the interviewers were blinded from the assignment of the interviewee to the Control or intervention group. After removing invalid numbers and duplicates, a total of 3024 beneficiaries across the two groups were called. Of these, 2555 (84.49\%) beneficiaries or related persons responded to the call. The interview was completed with 1554 respondents (51.39\%). The remaining were not interviewed due to unavailability of the mother, loss of pregnancy, or invalid registration details.  

This study showed significant improvement in scores for three key questions:
\begin{enumerate}
    \item ``Are you still taking iron pills after delivery?"
    \item ``Are you still taking calcium pills after delivery?"
    \item ``What was the baby’s weight at birth?", there is statistically significant improvement in score in the intervention group. 
\end{enumerate}

Table \ref{tab:study2} shows the scores and p-value for the three questions across control and intervention groups. 
The improved understanding and continued use of postnatal iron and calcium supplements for mothers in the intervention group is especially notable. This is because adherence after delivery is typically low due to limited counselling and poor follow-up at the health centres after delivery. Better awareness is crucial for maternal health as calcium helps restore bone mineral density depleted during pregnancy and lactation, while adequate iron prevents postpartum iron deficiency, which can impair maternal cognition and negatively affect infant growth.
Further, the knowledge of the baby's birth weight reflects whether the mother has tracked a key health indicator from birth. Birth weight is an important marker of neonatal health, guiding necessary home-care practices and follow-ups, especially for low-birth-weight infants. Accurate recall supports effective monitoring by health workers and aligns with WHO recommendations.

Overall, the higher scores in Table \ref{tab:study2} indicate stronger understanding and healthier practices related to micronutrient supplementation and newborn health.

\begin{table}[h!]
\centering
\resizebox{0.99\textwidth}{!}{
\renewcommand{\arraystretch}{1.3} 
\begin{tabular}{p{6cm} c c c c}
\hline
\textbf{Question} & \textbf{Correct Answer} & \textbf{Control Score} & \textbf{Intervention Score} & \textbf{p-Value} \\
\hline
Are you currently taking iron tablets after delivery? 
& Yes 
& \(0.234 \pm 0.019\) 
& \(0.283 \pm 0.021\) 
& 0.098111 \\
\hline
Are you still taking calcium pills after delivery? 
& Yes 
& \(0.244 \pm 0.019\) 
& \(0.306 \pm 0.022\) 
& 0.041262 \\
\hline
If yes, what was the baby’s weight at birth? (Follow-up to knowing weight) 
& \begin{tabular}[c]{@{}l@{}}`It was 2.5' or \\ `It was more than 2.5'\end{tabular}
& \(0.766 \pm 0.019\) 
& \(0.836 \pm 0.018\) 
& 0.007991 \\
\hline
\end{tabular}
}
\caption{Comparison of Control and Intervention Scores for Post-Delivery Health Questions}
\label{tab:study2}
\end{table}

\section{The SAHELI Deployment Architecture}
\begin{figure}
    \centering
    \includegraphics[width=0.9\linewidth]{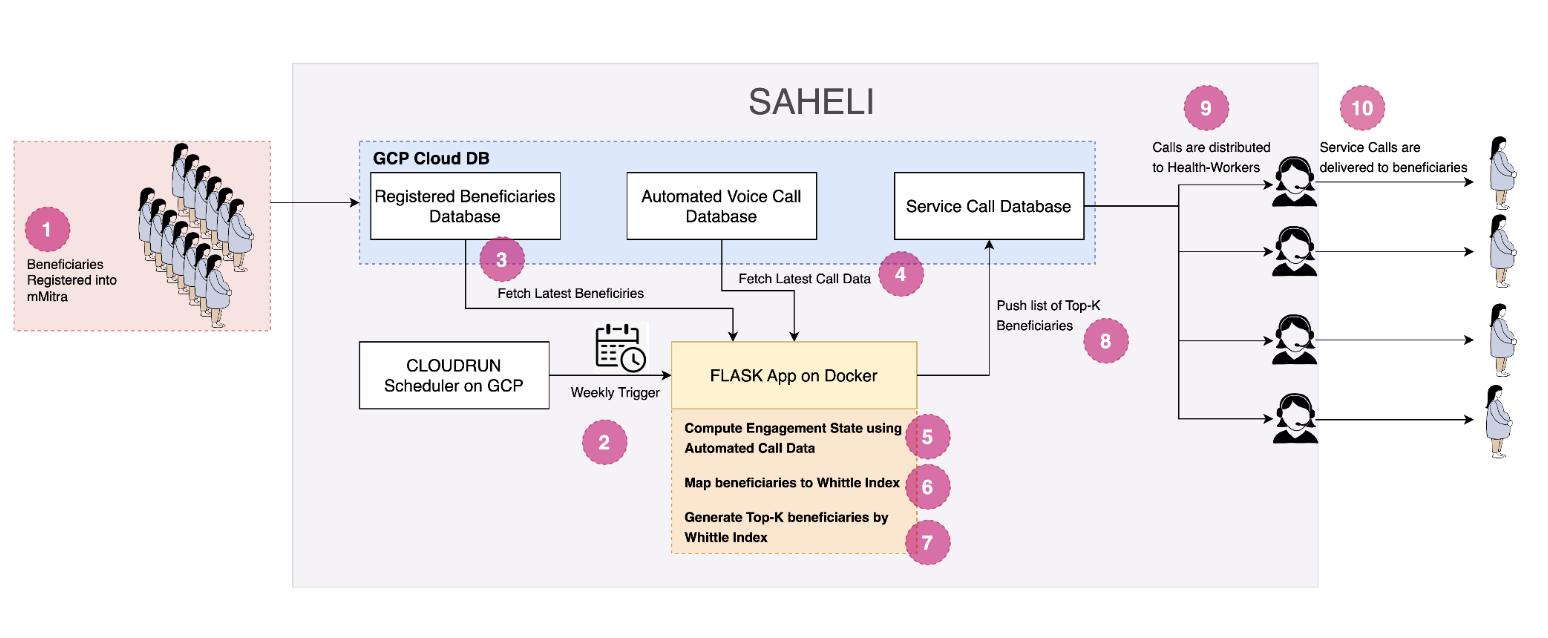}
    \caption{Pipeline of deployed system for scheduling live service calls. Reproduced from \cite{verma2022saheli}}
    \label{fig:saheli}
\end{figure}

[arch]
Deploying a complex optimization system like SAHELI within the operational context of a large NGO needs several careful design choices.  Figure \ref{fig:saheli} shows an outline of all the interactions within SAHELI’s ecosystem. Health workers periodically register beneficiaries through door-to-door visits or at hospitals (step 1). Socio-demographic information such as age, language, income range, and gestational age is entered into ARMMAN’s database (step 3). Automated voice messages tailored to gestational age are then delivered via a telecom provider (step 4), and metadata from these calls (e.g., duration, failure reason) is stored in ARMMAN’s database.

As engagement with voice messages declines over time, ARMMAN initiates live service calls to encourage continued participation (step 10). The AI pipeline predicts which beneficiaries would benefit most from receiving a service call, generating a prioritised list at the start of each week. This list is automatically distributed across health workers as shown in steps 2–9 of Figure \ref{fig:saheli}.


Step 8 in Figure \ref{fig:saheli} corresponds to generating the weekly intervention list, which is ingested into ARMMAN’s cloud databases. These databases support the client mobile application, which distributes scheduled service calls to health workers based on their weekly availability. SAHELI thus streamlines the deployment workflow into a single automated pipeline, making the process efficient, cost-effective, and easier to maintain and debug.

\section{Discussion and Key Lessons from Deployment}[lessons]
The journey from Restless Bandit theory to scaled deployment of the SAHELI system resulted in several critical lessons for the application of AI in social good domains:

\begin{enumerate}
    \item \textbf{Prioritise Decision Quality Over Predictive Accuracy:} The experience with Decision-Focused Learning (DFL) confirmed that in resource allocation, a model that is slightly inaccurate but leads to a superior final decision is far more valuable than a highly accurate prediction model that fails to align with the downstream optimisation task.
    \item \textbf{The Necessity of Resource-Aware Modeling:} The RMAB framework itself, by explicitly modelling the budget constraint, forces the solution to be resource-aware. Many social good problems are inherently resource-constrained; therefore, models like RMABs that are purpose-built for scarcity are generally more appropriate than models that assume unlimited resources.
    \item \textbf{Bridging the Gap to Behavioural Change:} The most recent study demonstrated the immense importance of tracking impact beyond easily measurable intermediate metrics (like listenership). The ultimate success of an mHealth intervention lies in improved health behaviours.
    \item \textbf{AI as an Augmentation, Not a Replacement:} The SAHELI system did not replace health workers; it improved their effectiveness. The human-AI collaboration model is highly successful: the AI provides the optimal schedule, and the human provides the empathy, cultural sensitivity, and effective counsel.
    \item \textbf{End-to-end integration}: 
    Testing of our application required our NGO partner to be equally involved in the validation of SAHELI’s outputs.
    Our experiences uncovering issues in the end-to-end pipeline led to improved communication practices, better documentation and tighter test goals. Social good applications like SAHELI have real-world consequences for beneficiaries in underserved communities, and it is critical that there be a real partnership for testing and integration.
\end{enumerate}

\section{Conclusion}[c]

The application of Restless Multi-Armed Bandits (RMABs) to the problem of optimising health worker service calls has proven to be a significant advancement in the efficiency and impact of mobile health programs. The SAHELI system, built upon this foundation and strengthened by innovations like Decision-Focused Learning, represents the successful operationalisation of AI techniques for a fundamental global health challenge.

The results from rigorous field studies demonstrate a multi-layered success: operational efficiency was maximised by dramatically increasing listenership and engagement rates within the program’s budget. Crucially, the latest research confirmed that this increased engagement translates directly into statistically significant improvements in maternal health behaviours.

Research and depolyment of SAHELI has led to a number of additional efforts in terms of both research and applications of restless bandits. Recent work has focused on using restless bandits for applications in maternal and child health in other countries, such as Uganda Uganda \cite{boehmer2025optimizing}, as well as in advancing the state of the art in restless bandits, such as using LLMs to enable easy steering of restless bandit policies by the end users \cite{behari2024decision}, and building a foundation model for restless bandits \cite{zhao2023towards}.

This body of work offers a compelling case for the role of sequential decision-making AI in solving resource-constrained societal problems. Moving forward, the blueprint established by SAHELI can be extended to numerous other domains requiring targeted, time-sensitive interventions under a limited budget, from wildlife anti-poaching and food security outreach to social service delivery. The future of AI for social good thus lies not just in predicting outcomes, but in optimising the allocation of human effort to realise the greatest positive impact.

\nocite{mate2022field,verma2022saheli,wang2020scalable,verma2023restless,dasgupta2025beyond}

\bibliographystyle{plain}
\bibliography{biblio}

\printindex
\end{document}